%% file: main.tex
\documentclass{article}

\usepackage{spconf,amsmath,graphicx}

\usepackage[utf8]{inputenc} 
\usepackage[T1]{fontenc}    
\usepackage{hyperref}       
\usepackage{url}            
\usepackage{booktabs}       
\usepackage{amsfonts}       
\usepackage{nicefrac}       
\usepackage{microtype}      
\usepackage[backend=biber,style=ieee,maxbibnames=1,minbibnames=1,mincitenames=1,maxcitenames=2,doi=false,isbn=false,url=false]{biblatex}
\usepackage{multirow}
\usepackage{siunitx} 

\title{From Visual to Acoustic Question Answering}

\name{Jerome Abdelnour$^{\dagger}$ \qquad Giampiero Salvi$^{\star}$ \qquad Jean Rouat$^{\dagger}$
\thanks{Acknowledgements to NVIDIA Corporation for GPU donations, to the Google Cloud Platform research credits program for computational resources. CHIST-ERA IGLU project, CRSNG scholarships \& Michael-Smith, UdeS.}
}

\address{$^{\dagger}$ NECOTIS, Department of Electrical and Computer Engineering\\
Sherbrooke University, Sherbrooke, J1K 2R1 \\
\texttt{\{Jerome.Abdelnour,Jean.Rouat\}@usherbrooke.ca} \\
    $^{\star}$ KTH Royal Institute of Technology\\
    School of Electrical Engineering and Computer Science\\
    Stockholm, Sweden \\
    \texttt{giampi@kth.se}
    }

\begin{document}
\maketitle


\begin{abstract}
We introduce the new task of Acoustic Question Answering (AQA) to promote research in acoustic reasoning.
The AQA task consists of analyzing an acoustic scene composed by a combination of elementary sounds and answering questions that relate the position and properties of these sounds.
The kind of relational questions asked, require that the models perform non-trivial reasoning in order to answer correctly.
Although similar problems have been extensively studied in the domain of visual reasoning, we are not aware of any previous studies addressing the problem in the acoustic domain.
We propose a method for generating the acoustic scenes from elementary sounds and a number of relevant questions for each scene using templates.
We also present preliminary results obtained with two models (FiLM and MAC) that have been shown to work for visual reasoning.
\end{abstract}


\input{introduction}

\input{data}
\input{method}
\input{experiments}
\input{results}
\input{conclusions}
\printbibliography

\end{document}

%% file: introduction.tex
\vspace{-3mm}
\section{Introduction and Related Work}
\subsection{Question Answering (QA)}
%
Question answering (QA) tasks are examples of constrained and limited scenarios for research in reasoning.
The agent's task in QA is to answer questions based on context.
Text-based QA use text corpora as context~\cite{voorhees1999trec, voorhees2000building, soubbotin2001patterns, hovy2000question, iyyer2014neural,ravichandran2002learning}. 
In visual question answering (VQA) the questions are related to a scene depicted in still images~\cite{johnson2017clevr,antol2015vqa,zhu2016visual7w,gao2015you, agrawal2016analyzing, zhang2016yin, geman2015visual, iyyer2014neural,ravichandran2002learning}.
Finally, video question answering attempts to use both the visual and acoustic information in video material as context~\cite{cao2005automated,chua2003question,yang2003videoqa,kim2017deepstory,tapaswi2016movieqa,wu2008robust}.
In the last case, however, the acoustic information is usually expressed in text form, either with manual transcriptions (e.g. subtitles) or by automatic speech recognition, and is limited to linguistic information~\cite{ZhangEtAl2017VQAwithspeech}.

In most studies, reasoning is supported by spatial and symbolic representations in the visual domain~\cite{chang1996symbolic,moktefi2013visual}. But reasoning and logic relationships are not attached and bounded exclusively to visual representations. They can be studied via representations of sounds~\cite{champagne2015}. This is of particular interest for research in artificial intelligence, and  has also implications in real world applications~\cite{champagne2018}.

The audio modality comprises important information that has not yet been exploited in QA reasoning. Audio allows QA systems to answer relevant questions more accurately, or even to answer questions that are not approachable from the visual domain alone.
In \cite{PieropanEtAl2014IROS}, e.g., audio was used in combination with video and depth information to recognize human activities.
The noise produced by a coffee machine can be used to know when the delivery of coffee is completed, to make another example.
Detection of abnormalities in machinery where the moving parts are hidden, or the detection of threatening or hazardous events are other examples of the importance of the audio information for QA.

Another interesting aspect of audio is that structures in an audio scene are organized through time. This is not covered in VQA in which structures of visual scenes are stationary and statics.  Audio scene analysis requires a memorization of time events and of their temporal structures. It requires longer-term memories and therefore opens the door to the development of new research tasks that can not be covered in current VQA research.
%
%
%

\subsection{Visual Question Answering (VQA)}
Several works have been proposed to build data for VQA and make them available to the scientific community~\cite{johnson2017clevr,antol2015vqa,zhu2016visual7w,gao2015you}.~\citeauthor{agrawal2016analyzing}~\cite{agrawal2016analyzing} noted that the type of  questions has a huge impact on the results of neural network based systems. This motivated researches to reduce the bias in VQA datasets~\cite{zhang2016yin, geman2015visual, johnson2017clevr}. The complexity around gathering good labeled data induced some authors \cite{zhang2016yin, geman2015visual} to constrain their work to yes/no questions. The recent work by \citeauthor{johnson2017clevr} \cite{johnson2017clevr} propose to create synthetic data for visual scenes in such way that none of the questions contains hints about the answer. Furthermore, the CLEVR database they created is extensively used to evaluate neural networks for VQA applications. The FiLM~\cite{perez2017film} and MAC~\cite{hudson2018compositional} neural networks have been tested on CLEVR. To generate the questions, they first build a semantic representation that describes the reasoning steps needed to answer the question. This gives them complete control over the labelling process and a better understanding of the semantic meaning of the questions. They leverage this ability to reduce the bias in the synthesized data.

\subsection{From Visual to Acoustic Question Answering}
We are not aware of previous research on applying question answering tasks to audio data. We therefore propose to introduce the new task of acoustic question answering (AQA). In this task the agent's goal is to answer questions related to \emph{acoustic scenes} composed by a number of \emph{elementary sounds}. The elementary sounds may be composed in sequences or mixed together. The questions relate the properties of the elementary sounds and their relative and absolute position in the scene.

The creation of audio databases for AQA applications is not trivial and several strategies can be adopted. Each strategy might introduce bias in the questions, in the answers and in the performance of the systems we want to evaluate.
In this study, in order to generate relevant questions and answers for each acoustic scene, we follow the same semantic representation through functional programs as proposed in~\cite{johnson2017clevr,johnson2017inferring} and described in~\cite{CleararXiv2018}.

Our research is part of a larger project in which we want  to study how a deep neural network, first trained on artificial data would transfer later on real-life natural scenes of sounds in the context of Acoustic Question Answering tasks. We, therefore, created an artificial database~\cite{CleararXiv2018} and report AQA results with the FiLM~\cite{perez2017film} and MAC~\cite{hudson2018compositional} neural networks.
These methods were initially designed by their respective authors for VQA tasks. Here we exploit them for the AQA task investigating their strengths and limitations when applied to audio. We modified the question generation to take the sequential nature of audio data into account. For this study we limit the acoustic scenes to be of equal length to allow the use of visual models. However, our data generation method allows us to produce more challenging data in the future, creating the opportunity to test models that were specifically designed for audio processing.
\subsection{Contributions}
This work is a first proposal for Compositional Language and Elementary Acoustic Reasoning in the context of Audio Question Answering.
We provide the software for an AQA task that comprises questions in relation with properties of the sounds in each scene. We provide as well a version of the first dataset that is based on musical instrument sounds.
We report experimental results using the FiLM and MAC architectures and analyze the results. 

%% file: data.tex
\section{Dataset} \label{sec:dataset}
To promote research in acoustic reasoning, we propose a data generation paradigm~\cite{CleararXiv2018} for AQA inspired by~\cite{johnson2017clevr} for VQA. The process generates a number of \emph{acoustic scenes}, of relevant questions for each scene, and the corresponding answer.


\newcommand{\emphPH}[1]{\textit{\textbf{#1}}}

\begin{table*}
    \centering
    \footnotesize
    \begin{tabular}{lp{0.48\textwidth}p{0.3\textwidth}c}
    \hline\hline
    Question type & Example & Possible Answers & \# \\
    \hline
    Yes/No       & Is there an equal number of \emphPH{loud} \emphPH{cello} sounds and \emphPH{quiet} \emphPH{clarinet} sounds? & yes, no & 2\\
    Note         & What is the note played by the \emphPH{flute} that is \emphPH{after} the \emphPH{loud} \emphPH{bright} \emphPH{D} note? & A, A\#, B, C, C\#, D, D\#, E, F, F\#, G, G\# & 12\\
    Instrument   & What instrument plays a \emphPH{dark} \emphPH{quiet} sound in the \emphPH{end} of the scene? & cello, clarinet, flute, trumpet, violin & 5 \\
    Brightness & What is the brightness of the \emphPH{first} \emphPH{clarinet} sound? & bright, dark & 2 \\
    Loudness & What is the loudness of the \emphPH{violin} playing after the \emphPH{third} \emphPH{trumpet}? & quiet, loud & 2 \\
    Counting     & How many other sounds have the same brightness as the \emphPH{third} \emphPH{violin}? & 0--10 & 11 \\
    Absolute Pos. & What is the position of the \emphPH{A\#} note playing after the \emphPH{bright} \emphPH{B} note? & \multirow{2}{*}{$\Big\}$ first--tenth} & \multirow{2}{*}{10} \\
    Relative Pos. & Among the \emphPH{trumpet} sounds which one is a \emphPH{F}? &  &  \\
    Global Pos. & In what part of the scene is the \emphPH{clarinet} playing a \emphPH{G} note that is \emphPH{before} the \emphPH{third} \emphPH{violin} sound? & beginning, middle, end (of the scene) & 3 \\
    \hline
    Total & & & 47 \\
    \hline\hline
    \end{tabular}
    \caption{Types of questions with examples and possible answers. The variable parts of each question is emphasized in bold italics. In the data set many variants of questions are included for each question type, depending on the kind of relations the question implies. The number of possible answers is also reported in the last column. Each possible answer is modelled by one output node in the neural network. Note that for absolute and relative positions, the same nodes are used with different meanings: in the first case we enumerate all sounds, in the second case, only the sounds played by a specific instrument.}
    \label{tab:types_of_question}
\end{table*}

\subsection{Scenes and Elementary Sounds} \label{sec:scenesAndSounds}
In this first version of the dataset, sounds are taken from a family of  real musical instruments:
cello, clarinet, flute, trumpet and violin.
Future versions may include natural sounds like speech, animals and environmental sounds.
Each elementary sound is described by an n-tuple [instrument family, brightness, loudness, musical note, absolute position, global position, relative position].
See Table~\ref{tab:types_of_question} for details on attributes values.
A various number of scenes have been created: from a minimum of 1 000 to a maximum of 50 000 scenes.
Once an auditory scene has been created it is filtered to simulate room reverberation using SoX\footnote{http://sox.sourceforge.net/sox.html}. For each scene, a different random room reverberation time is chosen, uniformly sampled in the interval [50ms, 400ms].

\subsection{Questions} \label{sec:questions}

%
Questions are structured in a logical tree expressing the hierarchy of the reasoning steps needed to answer. 
%
A function catalog 
is used 
to implement the relationships between the objects of the scene and the attributes of the objects~\cite{johnson2017clevr,CleararXiv2018}.
%
For example, one text template for the question “Is the cello as loud as the flute?” would be: “Is <I1> as loud as <I2> ?” where <I1> is an instrument's name and <I2> another name of instrument. Table~\ref{tab:types_of_question} gives examples of questions.
A template can be instantiated using a vast amount of element combinations. 
Not all of them generate valid questions. For example, the question “What is the position of the violin playing after the trumpet?” would be ill-posed if there are more than 1 violin playing after the trumpet. The same question would be considered degenerated if there is only one violin sound in the scene. It could be answered without taking into account the relation “after the trumpet”. A validation process~\cite{johnson2017clevr,CleararXiv2018} is responsible for the rejection of both ill-posed and degenerated questions during the generation phase.
A minimum of 20 000  and a maximum of 4 000 000 questions have been created. For more details, please refer to~\cite{CleararXiv2018}.

%% file: method.tex
\section{Methods}
We test two methods, FiLM~\cite{perez2017film} and MAC~\cite{hudson2018compositional} that proved to be effective in solving CLEVR. 
In order to do this, we need to make the simplifying assumption that an acoustic scene can be represented as a fixed resolution image. This is achieved in the current version of the data set, by creating scenes of the same duration. Future versions of the data will include scenes of varying lengths, thus requiring models that are more specific to audio processing.

Given our assumption, several options are available in this respect, e.g. linear frequency and Mel frequency spectrogram, Mel filterbank outputs, outputs of the gammatone filters, or Mel frequency cepstrum coefficients. Here we consider linear frequency spectrograms as representation of each scene. All the models we consider are based on convolutional neural networks. Although \cite{DielemanAndSchrauwen2014EndToEndAudio} proved that convolutional filters operating on raw audio samples tend to correspond to a logarithmic frequency response, there was no clear advantage using the logarthmic scale when applying the convolutions to the specrogram direclty \cite{Huzaifah2017LinearVsMel}. However, because using Mel filterbank is common practice in audio (e.g. \cite{pons2016experimenting}), we will in the future compare linear and Mel frequency scales.

Similarly to \cite{perez2017film} and \cite{hudson2018compositional}, we process the input images through the Resnet101~\cite{he2016deep} feature extractor trained on ImageNet~\cite{russakovsky2015imagenet}. We use the layer {\footnotesize\texttt{block3/unit\textunderscore22/bottleneck\textunderscore v1}} of Resnet101 which outputs 1024 feature maps of size $14\times 14$ for each input image. This representation is then fed as input either to the FiLM or the MAC network described below.
\subsection{FiLM}
FiLM~\cite{perez2017film} is a neural network inspired by the Conditional Batch Normalization architectures~\cite{IoffeAndSzegedy2015ICMLConditionalBatchNormalization}. The network takes as input an image (or image features) and a text-based question and then predicts the correct answer to the question given the scene. The visual input is processed by a sequence of convolutional layers that are linearly modulated by specific layers that are also called ``FiLM''. This modulation layer focuses attention on the relevant parts of the image given the question. The FiLM layer is driven by a number of gated recurrent units (GRUs) that process the text-based questions.
The combination of convolutional and FiLM layers is referred to as ResBlock in the model. A number of ResBlocks can be stacked together to increase the depth of the model. Finally, a softmax classifier predicts the most likely answer given the question and the scene.
This model was the state of the art for the CLEVR dataset at the time of its publication achieving 97.84\% of accuracy on the task. We used this model without modifications.

\subsection{MAC network}
The Memory Attention Composition (MAC) network~\cite{hudson2018compositional} is a question answering system aimed at multi-step reasoning. It uses a knowledge base and a text input that describes the task to be accomplished on the knowledge base. In our case, the knowledge base corresponds to an acoustic scene and the text input to the question. As with the FiLM model, visual features are extracted with Resnet101 and the question goes through a bidirectional LSTM network comprising 300 hidden units to create the vectorial representation of the question. The inference is executed by a series of MAC cells with recurrent connections between them. Each MAC cell comprises 3 gates (control, read and write) that dictate which information should be pushed to the internal memory of the network. The way the gates are designed force the MAC cells to become specialized into a specific reasoning function during training. The output of the final MAC cell is fed to a softmax layer that produces the distribution over the possible answers.
We used the same MAC architecture that achieves an averaged accuracy of 98.9\% on the CLEVR task.

\subsection{Baselines}
Because this is the first time that the CLEAR AQA task is proposed, there are no published previous results for reference. In order to give a measure of the complexity of the task, we report the chance level for the classification task (assuming uniform distribution of classes) and the majority class accuracy for the test sets.

%% file: experiments.tex
\begin{table}
    \centering
\resizebox{\columnwidth}{!}{
    \begin{tabular}{|ll|cc|ccc|}
        \hline\hline
        \multicolumn{2}{|c|}{Model} & \multicolumn{2}{c|}{Training data} & \multicolumn{3}{c|}{Accuracy (\%)} \\
        Name & Complexity & \# scenes & \# questions & Train & Eval & Test \\
        \hline
        
        FiLM & 4 ResBlocks & 700   & 14000   & 98.2 & 68.1 & 41.5 \\
        FiLM & 4 ResBlocks & 700   & 28000   & 97.2 & 75.8 & 42.8 \\
        FiLM & 4 ResBlocks & 7000  & 140000  & 94.6 & 81.4 & 51.2 \\
        FiLM & 2 ResBlocks & 7000  & 280000  & 97.4 & 87.2 & 52.9 \\
        FiLM & 4 ResBlocks & 7000  & 280000  & 97.4 & 89.4 & 58.0 \\
        FiLM & 4 ResBlocks & 35000 & 700000  & 97.9 & 94.6 & 89.0 \\
        FiLM & 4 ResBlocks & 35000 & 1400000 & 97.9 & 96.4 & 90.3 \\
        \hline
        MAC  & 4 MacCells  & 7000  & 140000  & 92.7 & 88.9 & 44.2$^*$ \\
        MAC  & 8 MacCells  & 7000  & 140000  & 94.8 & 88.0 & 44.8$^*$\\
        \hline\hline
    \end{tabular}
    }
    {\footnotesize $^*$) the MAC test results were not complete at the time of submission, we report here results on a smaller test set but will update them before publication.}

    \caption{Summary of the results. All test results are obtained on the same test set of 7500 scenes and 300000 questions and answers.}
    \label{tab:results}
\end{table}

\vspace{-4mm}
\section{Experiments} \label{sec:experiments}

\vspace{-2mm}
\subsection{Sound and data preprocessing}
Each scene is about 50 seconds long and is sampled at 48kHz, the spectrogram is computed with window size of 1024 samples (about 21 msec) and Hanning window, and time shifts of 512 samples (about 10.7 msec). The spectrogram is then considered as an image and resampled to $480\times 320$ pixels to fit the input layer of the Resnet101 network. 

The number of scenes in the data is varied from 1000 to 10000 and 50000 scenes. Moreover, for each acoustic scene we generated either 20 or 40 questions and answers per scene resulting in six distinct sets of data. From each of these sets, we split the data into training, validation and test with the sizes of 70\%, 15\% and 15\% of the total, respectively. No scene is shared between training, validation and test set.
In order to facilitate the comparison between the performance of each model, the results are reported for the same test set which includes 7500 scenes and 300000 questions and answers.

\vspace{-2mm}
\subsection{Training and evaluation}
The FiLM model was trained for 25 epochs with cross entropy loss using the Adam optimizer with \num{1e-5} weight decay, \num{3e-4} learning rate and 512 batch size. Similarly, the MAC network was trained for 25 epochs with Adam and \num{1e-4} learning rate and batch size of 64.
The results are reported in terms of accuracy, that is the percentage of correct answers on the total. We also detail the accuracy depending on the question type and answer (see Table~\ref{tab:types_of_question})

%% file: results.tex

\section{Results}
\label{sec:results}
The evaluation results are reported in Table~\ref{tab:results} for the different models, model complexities and amount of training data. The table reports results each training and validation sets and for a test set that is common to all models and includes 7500 scenes and 300000 question and answers. The chance level for the classification problem is $1/47 = 2.1\%$. Given the distribution of answers in the training evaluation and and test set, the majority class accuracy (accuracy choosing always the most common class) is 7.6\%. The Table shows that there is a fair amount of over-training when the networks are trained on smaller data sets. The difference in results between the evaluation and test set in those cases might be due to the different distribution of questions and answers in those sets. Increasing the number of ResBlocks in the FiLM network from 2 to 4 gives no improvement on the training set, but a large improvement on the test set. For the MAC model, instead, we see an improvement on the training set when using 8 MacCells instead of 4, but a slight reduction and a small improvements for the evaluation and test sets.

The best results of 90.3\% accuracy is obtained with FiLM network with 4 ResBlocks trained on the largest data set.

If we analyze the accuracy for each question type reported in Table~\ref{tab:types_of_question}, we observe that the models trained on 35k scenes have the same accuracy across all question types while the other models show a clear dependency on question type. In particular the note, and relative positions are the most difficult questions (accuracies around 25--45\%), whereas the loudness and brightness are the easiest questions (accuracies around 60--70\%).

%% file: conclusions.tex
\section{Conclusion}
We propose a framework for Acoustic Question Answering problems, a method to create  auditory scenes in that context
and report results with 2 neural networks that were initially designed for Visual Question Answering tasks.
While FiLM and MAC reach respectively accuracy of 97.8\% and 98.9\% on the CLEVR visual task, we get a maximum of 90\% on the CLEAR acoustical task for the network trained on the largest data set. It is still impressive to obtain such good results without any changes in the architectures of the networks. On the other hand, with a smaller and more realistic number of acoustical scenes, results are around 40\%. A research avenue would be the use of artificial large dataset to train and then study and develop new means to transfer to real life situations. 

Much improvements can be made to this initial CLEAR database (add more primary sounds, more natural and environmental sounds, speech, more variability in the scenes, overlap sounds, variable length scenes, etc.). The software and the code of this first version is available to the community for improvement and new research avenues.